%% file: paper.tex
\documentclass{article}
\usepackage[final]{corl_2021}
\usepackage[numbers]{natbib}
\usepackage{multicol}
\usepackage{wrapfig}
\usepackage{amsmath} % assumes amsmath package installed
\usepackage{amssymb}  % assumes amsmath package installed
\usepackage{siunitx} % for SI units
\usepackage{gensymb} % for degree command
\usepackage{booktabs}
\usepackage{multirow}
\usepackage{xcolor}
\usepackage{booktabs}
\usepackage{siunitx}
\usepackage{todonotes}
\usepackage{subcaption}
\usepackage{caption}
\usepackage{xspace}
\newcommand{\mypara}[1]{\par\vspace*{0.8mm}\noindent\textbf{{#1}}}

\usepackage{tikz}
\usetikzlibrary{arrows}
\usetikzlibrary{shapes}

\newcommand{\OURS}{FlingBot\xspace}

\title{
FlingBot: The Unreasonable Effectiveness of Dynamic Manipulation for Cloth Unfolding 
}

\author{
Huy Ha \qquad Shuran Song\\
Columbia University\\
\href{https://flingbot.cs.columbia.edu}{https://flingbot.cs.columbia.edu}
}

\begin{document}
\maketitle
\vspace{-5mm}
\input{text/abstract}
\vspace{-2mm}
\keywords{Dynamic manipulation, Cloth manipulation, Self-supervised learning}
\vspace{-3mm}
\input{text/intro}
\vspace{-3mm}
\input{text/related}
\vspace{-3mm}

\input{text/method}

\vspace{-3mm}

\input{text/eval}
\vspace{-3mm}
\input{text/conclusion}
\acknowledgments{
  The authors would like to thank Eric Cousineau, Benjamin Burchfiel Naveen Kuppuswamy, other researchers in Toyota Research Institute, Zhenjia Xu, and Cheng Chi for their helpful feedback and fruitful discussions.
  We would also like to thank Google for their donation of UR5 robot hardware.
  This work was supported in part by the Amazon Research Award and the National Science Foundation under CMMI-2037101.
}

\bibliography{paper}
\newpage
\input{text/supp}
\end{document}

%% file: text/abstract.tex
\begin{abstract}
    High-velocity dynamic actions (e.g., fling or throw) play a crucial role in our everyday interaction with deformable objects by improving our efficiency and effectively expanding our physical reach range.
    Yet, most prior works have tackled cloth manipulation using exclusively single-arm quasi-static actions, which requires a large number of interactions for challenging initial cloth configurations and strictly limits the maximum cloth size by the robot's reach range.
    In this work, we demonstrate the effectiveness of dynamic flinging actions for cloth unfolding with our proposed self-supervised learning framework, FlingBot.
    Our approach learns how to unfold a piece of fabric from arbitrary initial configurations using a pick, stretch, and fling primitive for a dual-arm setup from visual observations.
    The final system achieves over 80\% coverage within 3 actions on novel cloths,  can unfold cloths larger than the system's reach range, and generalizes to T-shirts despite being trained on only rectangular cloths.
    We also finetuned FlingBot on a real-world dual-arm robot platform, where it increased the cloth coverage over 4 times more than the quasi-static baseline did.
    The simplicity of FlingBot combined with its superior performance over quasi-static baselines demonstrates the effectiveness of dynamic actions for deformable object manipulation.
\end{abstract}

%% file: text/intro.tex
\section{Introduction}
\vspace{-2mm}
\begin{wrapfigure}{r}{0.5\textwidth}
    \vspace{-6mm}
    \centering

    \includegraphics[width=\linewidth]{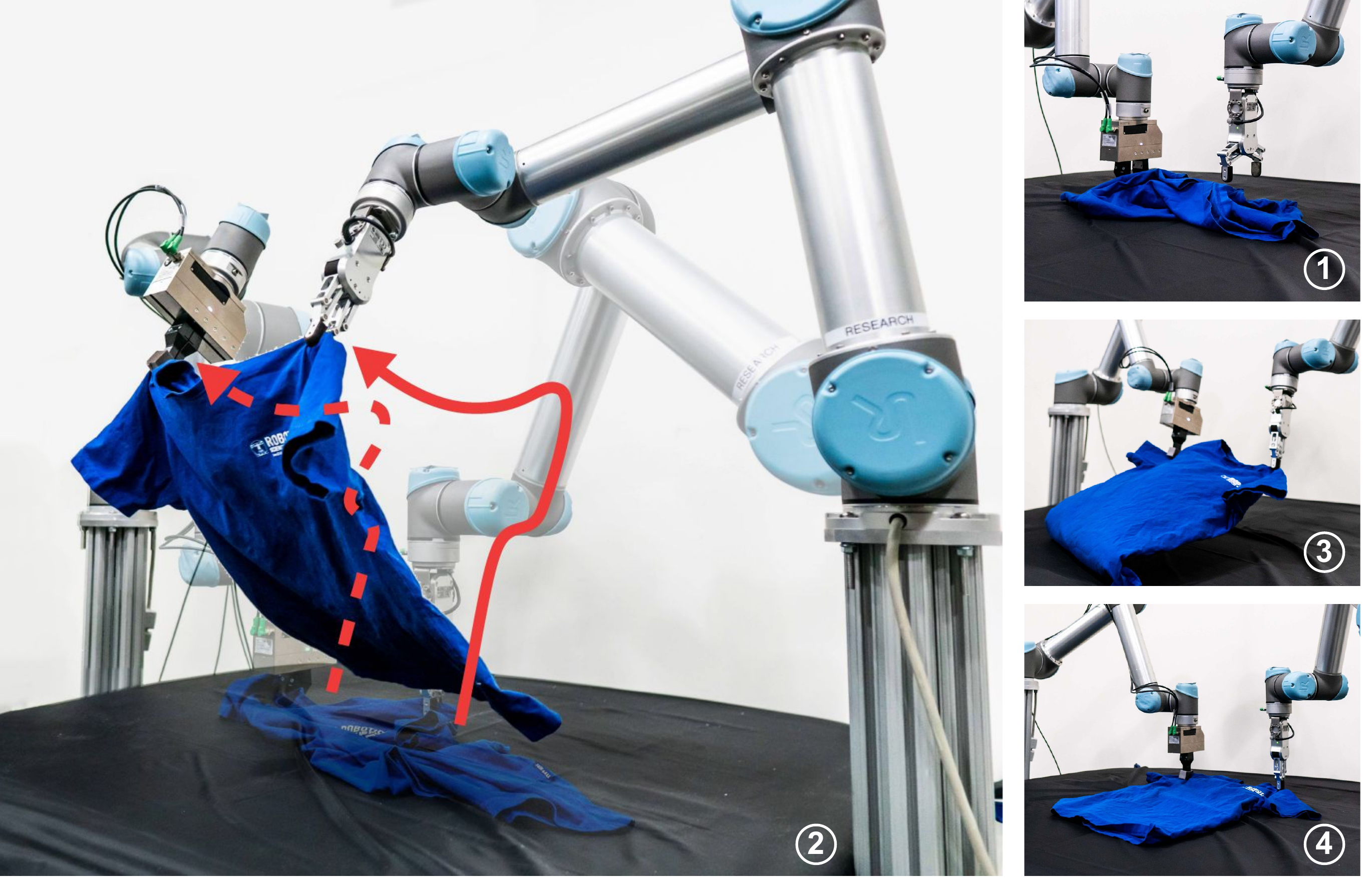}
    \vspace{-3mm}
    \caption{
        \textbf{Cloth unfolding with dynamic interactions.}
        Given a severely crumpled cloth, \OURS uses a high-speed fling to unfurl the cloth with as little as one interaction.
        In this paper, we demonstrate that such dynamic actions can efficiently unfold cloths, generalizable to different cloth types, and improve the effective reach range of the system.
    }
    \vspace{-4mm}
    \label{fig:teaser}
\end{wrapfigure}

High-velocity dynamic actions play a crucial role in our everyday interaction with deformable objects.
Making our beds in the morning is not effectively accomplished by picking up each corner of the blanket and placing them in the corresponding corners of the bed, one by one.
Instead, we are more likely to grasp the blanket with two hands, stretch it, and then unfurl it with a fling over the bed.
This fluid high-velocity flinging action is an example of dynamic manipulation~\cite{dynamic1993mason}, which is used to improve our physical reachability and action efficiency --  allowing us to unfold large crumpled cloths with as little as one interaction.

From goal-conditioned folding~\cite{lee2020learning} to fabric smoothing~\cite{seita2019imitation,wu2019withoutdemonstrations}, prior works have achieved success using exclusively single-arm quasi-static interactions (e.g., pick \& place) for cloth manipulation.
However, these approaches require a large number of interactions for challenging initial configurations (e.g., highly crumpled cloths) or rely on strong assumptions about the cloth (e.g., predefined keypoints).
Additionally, since the robot arm cannot manipulate the cloth at locations it can't reach, the maximum cloth size is greatly limited by the robot arm's reach range.

In this work, we focus on the task of cloth unfolding, a typical first step for many cloth manipulation tasks.
The goal of this task is to maximize the cloth's coverage through interactions to expose the key visual features for downstream perception and manipulation.
Therefore, an ideal cloth unfolding approach should be:
\begin{itemize}
    \item \textbf{Efficient}: the approach should reach a high coverage with a small number of actions from arbitrarily crumpled initial configurations.
    \item \textbf{Generalizable}: the algorithm should not rely on heuristics (e.g., grasp predefined key points, etc.).
          This is especially important when unfolding is only the initial stage of a cloth perception and manipulation pipeline, where key points are not visible or severely occluded, and when the system must handle cloth types unseen during training, which may not contain the predefined key points.
    \item \textbf{Flexible Beyond the Workspace}: the approach should work with cloths of different sizes, including large ones which lie outside the robot's physical reach range.
\end{itemize}

To achieve this goal, we present \OURS, a self-supervised algorithm that learns how to unfold cloths from arbitrary initial configurations using a pick, stretch, and fling primitive for a dual-arm setup.
At each time step, the policy predicts value maps from its visual observation and picks actions greedily with respect to its value maps.
To provide the supervision signal, the system computes the difference in coverage of the cloth before and after each action -- the delta-coverage -- from the visual input captured by a top-down camera.
\OURS achieves over 80\% coverage within 3 actions on novel cloths and increases the cloth's coverage by more than twice that of pick \& place and pick \& drag quasi-static baselines on rectangular cloths.
Our approach is flexible to large cloths whose dimensions exceed the robot arm's reach ranges and generalizes to T-shirts despite being trained on rectangular cloths.
We fine-tune our approach in the real world, where, on average, it increased the cloth coverage over 4 times more than the quasi-static pick \& place baseline did.
In summary:
\begin{itemize}
    \item Our main contribution is in demonstrating the effectiveness of dynamic manipulation for cloth unfolding through our self-supervised learning framework, \OURS.

    \item We propose a parameterization for the dual-arm grasp of our fling primitive, which enables the application of the simple yet effective single-arm grasping technique~\cite{zeng2018learning,zeng2019tossingbot,Wu_2020} to dual-arm grasping while satisfying safety constraints of dual-arm systems.
          The simplicity of FlingBot combined with its superior performance over quasi-static baselines further emphasize the effectiveness of dynamic actions for deformable object manipulation.

    \item We present a custom simulator\footnote{
Please visit \href{https://flingbot.cs.columbia.edu}{https://flingbot.cs.columbia.edu} for experiment videos, code, simulation environment, and data.} built on top of PyFlex~\cite{li2019learning,lin2020softgym}, a CUDA accelerated simulator, which supports the loading of arbitrarily shaped cloth meshes.
          We hope this open-source simulator expands cloth manipulation research to more complex cloth types.
\end{itemize}

%% file: text/related.tex
\section{Related Work}
\vspace{-2mm}
A convincing argument for the \emph{addition} of dynamic actions to previously exclusively quasi-static cloth manipulation pipelines would need to demonstrate their superior performance on a core cloth manipulation skill.
As a typical first step for many cloth manipulation tasks, cloth unfolding is a popular and important problem setting for studying deformable object manipulation.
The goal of cloth unfolding is to maximize the coverage of the cloth on the workspace, which exposes key visual features of the cloth for downstream applications.
However, achieving a fully unfolded cloth configuration from a severely crumpled initial configuration remains a challenging problem.
Additionally, doing so efficiently for many different types of cloths, some larger than the system's reach range, is extremely challenging.

\mypara{Quasi-static Cloth Manipulation with Expert Demonstrations.}
Prior works have explored using heuristics and identifying key points such as wrinkles~\cite{Sun2013AHA}, corners~\cite{willimon2011model,seita2018bedmaking,maitin2010cloth}, edges~\cite{triantafyllou2016geometric}, or combinations of them~\cite{yuba2017unfolding}, but fail to generalize to severely self-occluded cloth configurations, to when required key points aren't visible, or to non-square cloths.
Recent reinforcement learning approaches~\cite{seita2019imitation,ganapathi2020learning} relied on cloth unfolding expert demonstrations in quasi-static pick-and-drag action spaces,
While they sidestep the exploration problem, expert demonstrations can be sub-optimal and brittle if from hard-coded heuristics~\cite{seita2019imitation} or expensive if from humans~\cite{ganapathi2020learning}.

\mypara{Self-supervised Quasi-static Cloth Manipulation.}
Bypassing the dependency on an expert, self-supervised cloth manipulation has been demonstrated in unfolding with a factorized pick \& place action space by \citet{wu2019withoutdemonstrations} and goal conditioned folding using spatial action maps~\cite{Wu_2020,zeng2019tossingbot,zeng2018learning} by \citet{lee2020learning}.
However, all these approaches operate entirely in quasi-static action spaces.

\mypara{Dynamic Cloth Manipulation.}
In contrast to quasi-static manipulation, which are ``operations that can be analyzed using kinematics, static, and quasi-static forces (such as frictional forces at sliding contacts)''~\cite{dynamic1993mason}, dynamic manipulation involves operations whose analysis additionally requires ``forces of acceleration''.
Intuitively, dynamic manipulation (e.g., tossing~\cite{zeng2019tossingbot}) involve high-velocity actions which build up objects' momentum, such that the manipulated objects continue to move after the robot's end-effector stops.
Such dynamic manipulation results in an effective increase in reach range and a reduction in the number of actions to complete the task compared to exclusively quasi-static manipulation.
However, prior works on dynamic cloth manipulation have either relied on cloth's vertex positions~\cite{jangir2019dynamic} (so is only practical in simulation), a motion capturing system with markers combined with human demonstrations~\cite{balaguer2011combining}, or custom hardware~\cite{yamakawa2011dynamic,shibata2010robotic}.

While prior cloth unfolding works report high average final coverages from relative high average initial coverages, we emphasize that we consider severely crumpled initial cloth configurations, which are much more challenging yet still realistic (see section~\ref{sec:task_generation}).
Additionally, no prior works have considered unfolding cloths larger than the robot's reach range (indeed, \citet{ganapathi2020learning}'s workspace is significantly larger than their maximum cloth size due to physical reachability limitations).
Our algorithm achieves high performance in these challenging cases through self-supervised trial and error and learns directly from visual input without requiring expert demonstrations or ground truth state information.

%% file: text/method.tex
\section{Method}
\vspace{-2mm}
The goal of cloth unfolding is to manipulate a cloth from an arbitrarily crumpled initial state to a flattened state.
Concretely, this amounts to maximizing the cloth's coverage on the workspace surface.
Intuitively, dynamic actions have the potential to achieve high performance on cloth unfolding by appropriately making use of the cloth's mass in a high-velocity action to unfold the cloth (Sec.~\ref{sec:advantages}).

From a top-down RGB image of the workspace with the cloth, our policy decides the next fling action (Sec.~\ref{sec:motion_primitives}) by picking the highest value action which satisfies the system's constraints (Sec.~\ref{sec:constraint_action_primitives}).
It predicts the value of each action with a value network (Sec.~\ref{sec:learning_value_maps}) which is trained in a self-supervised manner to take actions that maximally increases the cloth's coverage.
The supervision signal is computed directly from the visual observation captured by the top-down camera.
After training in simulation, we finetune and evaluate the model in the real world (Sec.~\ref{sec:exp_setup}).

\vspace{-2mm}
\subsection{Advantages of the Fling Action Primitive}\label{sec:advantages}
\vspace{-2mm}

Quasi-static actions, such as pick \& place, rely on friction between the cloth and ground to stretch the cloth out.
The complexity of friction forces between the workspace and the (nonvisible, ground-facing) surfaces of the cloth means the cloth's final configuration may be difficult to predict from only visual observations, especially for novel cloth types or significantly different friction forces (i.e.: simulation v.s. real).
In addition, such systems can't manipulate points on the cloth to regions outside of their physical reach range, which limits their maximum cloth dimensions.
Meanwhile, dynamic actions largely rely on cloths' mass combined with a high-velocity throw to do most of its work.
Since a single dynamic motion primitive can effectively unfold many cloths, dynamic unfolding systems can learn a simpler policy which generalizes better to different cloth types.
In addition, they can reach higher coverages in smaller numbers of interactions and throw corners of cloths larger than the system's reach range, effectively expanding their physical reach range.

We propose to use a dual-arm pick, stretch, and fling primitive.
Here, a dual-arm system stretches the cloth between the arms, which unfolds it in one direction, and then flings the cloth, which makes use of the cloth's mass to unfold it in the other direction.
The combination of stretching and flinging ensures that, given two appropriate grasp points, our motion primitive should be sufficient for single step unfolding when possible.

\vspace{-2mm}
\subsection{Fling Action Primitive Definition}\label{sec:motion_primitives}
\begin{figure*}[t]
    \centering
    \includegraphics[width=\textwidth,trim=0 0 3 0, clip]{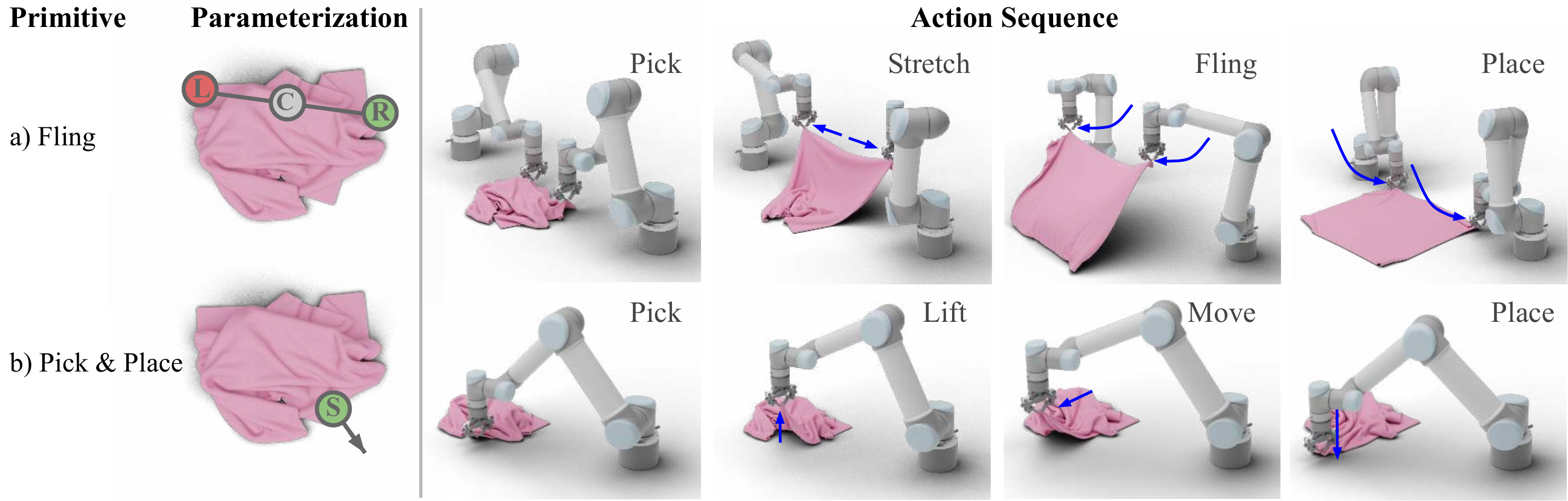}
    \vspace{-3mm}
    \caption{\textbf{Action Primitives.}
        The dynamic \textbf{Fling} primitive starts with a two-arm grasp at the left $L$ and right $R$ grasp locations with center point $C$, followed by a fixed stretch, fling, and place routine. A good fling could unfold a highly crumpled cloth in as little as a single step. In contrast, the quasi-static  \textbf{Pick \& Place} primitive, which grasps at the start location $S$, lifts, moves, then places at the end position specified by the arrow tip, requires many steps for such challenging cloth configurations. The quasi-static  \textbf{Pick \& Drag} primitive is just a Pick \& Place without the lift step.
    }
    \label{fig:primitives}
    \vspace{-5mm}
\end{figure*}

\vspace{-2mm}
To achieve an efficient, generalizable, and flexible cloth unfolding system, we argue that the system requires two arms operating in a dynamic action space.
To this end, we've designed a pick, stretch, and fling motion primitive for a dual-arm system, where each arm is placed on either side of the cloth workspace, as follows.
First, the arms perform a top-down pinch-grasp on the cloth at locations $L, R\in \mathbb{R}^3$.
Second, the arms lift the cloth to $0.30\si{\meter}$ and stretch the cloth taut in between them.
Third, the arms fling the cloth forward $0.70\si{\meter}$ at $1.4\si{\meter\per\second}$ then pull backwards $0.50\si{\meter}$ at $1.4\si{\meter\per\second}$, which swings the cloth forwards.
Finally, the arms place the cloth down to the workspace and release their grips.
The entire motion primitive is visualized in (Fig.~\ref{fig:primitives}a).

Instead of parameterizing and learning all steps of our primitive, we can fix the stretching step to always stretch the cloth as much as possible without tearing the cloth.
We also fix the fling speed and trajectory from the observation that the real world system could robustly unfold the cloth using a wide range of fling parameters (i.e.: fling height, fling speed) if given a good grasp
(\makeatletter\@ifundefined{r@fig:real_world_fling_robustness}{see supplementary materials}{Fig.~\ref{fig:real_world_fling_robustness}}\makeatother).
Thus, the problem of learning to pick, stretch, and fling reduces down to the problem of learning where to pick, which is parameterized with only two grasp locations, one for each arm.

\vspace{-2mm}
\subsection{Constraint Satisfying Fling Action Parameterization}\label{sec:constraint_action_primitives}
\vspace{-2mm}

A dual-arm grasp is parameterized by two points $L,R\in \mathbb{R}^3$, which denotes where the left and right arm should approach from a top-down grasp respectively, and requires 6 scalars.
Without loss of generality for the purposes of grasping from a top-down RGB-D input, the third dimension could be specified by depth information.
This reduces $L$ and $R$ to two points in $\mathbb{R}^2$, each representing pixels to grasp from the visual input and uses only 4 scalars in total.

However, to minimize collisions between two arms, we wish to impose a constraint that $L$ is always left of $R$, and vice versa \makeatletter\@ifundefined{r@fig:crossover_constraint}{}{(Fig~\ref{fig:crossover_constraint})}\makeatother.
Additionally, the grasp width (i.e., the length of the line $L-R$) must not exceed the physical limit of the system and be smaller than the minimum safe distance limit between the two arms \makeatletter\@ifundefined{r@fig:grasp_width_constraint}{}{(Fig.~\ref{fig:grasp_width_constraint})}\makeatother.
Directly using $L$ and $R$ will entangle these two contraints  \makeatletter\@ifundefined{r@sec:fig:grasp_width_constraint}{(see supplementary materials)}{}\makeatother, making them difficult to satisfy.
To make these constraints linear and independent, we propose an alternative 4-scalar parameterization, which consists of pixel position of the point $C \in \mathbb{R}^2$ at the center of the line $L-R$, an angle $\theta \in \mathbb{R}$ denoting the planar rotation of the line $L-R$, and a grasp width $w \in \mathbb{R}$ denoting the length of the line in pixel space.
To constrain $L$ to be on the left of $R$, we can constrain $\theta \in [-90^\circ,90^\circ]$, while $w$ can be directly constrained to appropriate system limits.

\vspace{-2mm}
\subsection{Learning Delta-Coverage Maximizing Fling Actions}\label{sec:learning_value_maps}
\vspace{-2mm}

The naive approach for learning the action parameters $\langle C_x, C_y, \theta, w\rangle$ by directly predicting these 4 scalars does not accommodate the best grasp's equivariances to the cloth's physical transformations.
However,
the learner should be equipped with the inductive bias that cloths in identical configurations have identical optimal grasp points for flinging relative to the cloth, regardless of the cloth's translation, rotation, and scale.

\input{text/spatial_action_maps_concise}

\begin{figure*} [t]
    \centering
    \includegraphics[width=\linewidth]{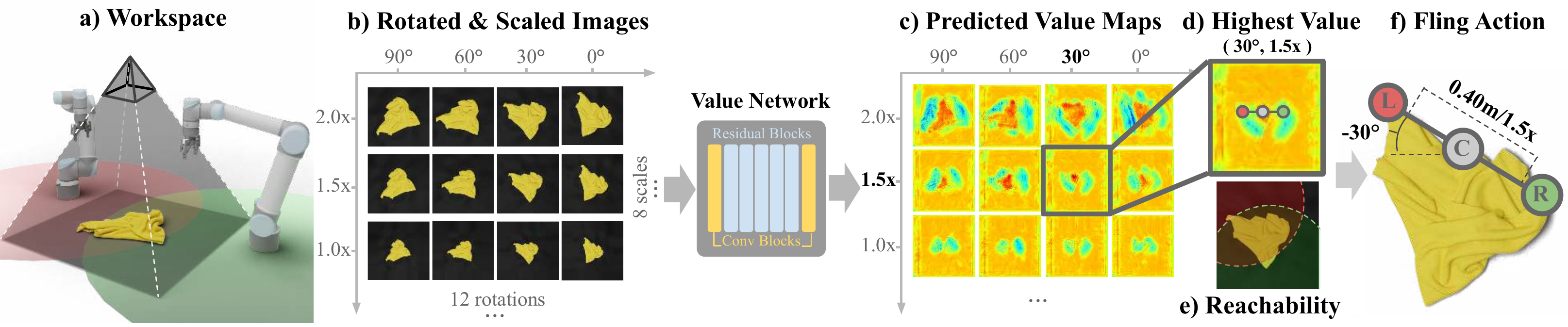}
    \caption{ \textbf{Method Overview. }
        From a top-down RGB image a), our policy evaluates a batch of different action rotations and scales by transforming the observation b) then predicting the corresponding batch of value maps c).
        The highest value action d), corresponding to the maximum value pixel which also satisfies the arms' reachability constraints e) is chosen.
        Finally, the chosen pixel's location and its observation's transformation is decoded into the fling action parameters (i.e., center point, distance, relative orientation between gripper). \vspace{-3mm}
    }
    \label{fig:approach}
    \vspace{-2mm}
\end{figure*}

\mypara{Self-Supervised Learning. }
The value network is trained end-to-end with self-supervised trials in simulation then finetuned in the real world.
In each training episode, a random task is sampled.
Each fling step is labeled with its normalized delta coverage, which is computed by counting cloth mask pixels from a top-down view, then dividing by the cloth mask pixel count of a flattened cloth state.
While our experiments have plain black workspace surface colors and non-textured colored cloths for easy masking, our approach can be combined with image segmentation or background subtraction approach to obtain the corresponding cloth mask to work with arbitrary cloth and workspace textures.
Note that this supervision signal (i.e., cloth mask) is only needed during training.

The policy interacts with the cloth until it reaches 10 timesteps or predicts grasps on the workspace. The latter stopping condition indicates that the policy does not expect any possible action to further improve the cloth coverage.
All policies are trained in simulation until convergence, which takes around 150,000 simulation steps, or 6 days on a machine with a GTX 1080 Ti.
The network is trained using the Adam optimizer with a learning rate of 1e-3 and a weight decay of 1e-6.
\vspace{-2mm}
\subsection{Experiment Setup}\label{sec:exp_setup}
\vspace{-2mm}

\mypara{Simulation Setup.}
Our custom simulation environment is built on top of the PyFleX~\cite{li2019learning} bindings to Nvidia FleX provided by SoftGym~\cite{lin2020softgym}.
On top of the functionality provided by PyFleX and SoftGym, our custom simulator\footnote{Our custom simulator is publicly accessible at \href{https://github.com/columbia-ai-robotics/flingbot}{https://github.com/columbia-ai-robotics/flingbot}} can load arbitrary cloth meshes, such as T-shirts, through the Python API.
While our simulation environment does not support loading URDFs of robots, we found it sufficient to represent only the arms' end effectors, and apply appropriate physical constraints to their locations.
The observations are rendered using Blender 3D where the cloth HSV color is sampled uniformly between $[0.0,1.0]$, $[0.0,1.0]$, and $[0.5,1.0]$ respectively, and the background is dark grey with a procedurally generated normal map to mimic the wrinkles on the real-world workspace surface.
We further apply brightness, contrast, and hue jittering on observations to help with the transfer to real.
We manually tuned our simulation fling speed to qualitatively match the real-world cloth dynamics (see \makeatletter\@ifundefined{r@sec:real_world_fling_speed}{supplementary materials }{Sec.~\ref{sec:real_world_fling_speed} }\makeatother
on simulation sensitivity and real world robustness to fling parameters).

\mypara{Real World Setup.}
Our real-world experiment setup consists of two UR5s, where one is equipped with a Schunk WSG50 and the other with an OnRobot RG2, facing each other and positioned $1.35\si{\meter}$ apart.
The top-down RGB-D image is captured with an Intel RealSense D415.
To help with pinch grasp success in real, we used a 2-inch rubber mat as the workspace and added sponge tips to the gripper fingers.
To mitigate the reliance on specialized and expensive force sensor, we chose to implement stretched cloth detection using only a second Intel RealSense D415 capturing a frontal view of the workspace.
By segmenting the cloth from the front camera RGBD view of the workspace, then checking whether the top edge of the cloth mask is flat as a proxy for the cloth being stretched, we can pull the arms' end effectors apart until the top of the cloth is no longer bent.

%% file: text/spatial_action_maps_concise.tex
To this end, we propose to use spatial action maps~\cite{zeng2018learning,zeng2019tossingbot,Wu_2020}.
By predicting grasp values with constant scale and rotation \emph{in pixel space} from the transformed images, spatial action maps can recover grasp values with varying scales and rotations \emph{in world space} by varying the transformation applied to the image.
Concretely, given a visual observation from the top-down view (Fig.~\ref{fig:approach}a), we generate a batch of rotated and scaled observations (Fig.~\ref{fig:approach}b) then predict the corresponding batch of dense value maps (Fig.~\ref{fig:approach}c).
Each pixel in each value map contains the value of the action parameterized by that pixel's location, giving $C$, and its observation's rotation and scaling, giving $\theta$ and $w$ respectively (Fig.~\ref{fig:approach}f).
The value network is supervised to regress each pixel in the value map to the ground truth delta-coverage -- the difference in coverage before and after the action.
Thus, by picking the grasping action with the highest value (Fig.~\ref{fig:approach}d), the system picks grasp points which it expects would lead to the greatest increase in cloth coverage.
By its architecture, the value network is invariant to translation.
By rotating and scaling its inputs, it is also invariant to rotation and scale.

In our approach, we use 12 rotations, which discretizes the range $[-90^\circ,90^\circ]$, and $8$ scale factors in the range $[1.00,2.75]$ at $0.25$ intervals.
Our value network is a fully convolutional neural network with nine residual blocks~\cite{he2016deep} and two convolutional layers in the first and last layer, and takes as input $64\times 64$ RGB images.
To evaluate actions within a useful range of scales, we crop the image such that the cloth takes up at most two-thirds of the image width and height before shrinking it by the inverse of the $8$ scale factors above.
We also filter out and reject all grasp point pairs which are out of reach for either arm.

%% file: text/eval.tex
\section{Evaluation}
\vspace{-2mm}
We design a series of experiments to evaluate the advantage of dynamic actions over quasi-static actions in the task of cloth unfolding.
The system's performance is evaluated on its efficiency (i.e., reaching high coverages with a small number of actions), generalization to unseen cloth types (i.e., flattening different types of T-shirts when only trained on rectangle cloths), and reach range (i.e., on cloths with dimensions larger than its reach range).
Finally, we evaluate the algorithm's performance on the real-world setup.
Please visit \href{https://flingbot.cs.columbia.edu}{https://flingbot.cs.columbia.edu} for experiment videos.

\vspace{-2mm}
\subsection{Metrics}\label{sec:metrics}
\vspace{-2mm}
The algorithm performance is measured by the final coverage, delta-coverage from initial coverage, and the number of interactions.
Final coverage measures the coverage at the end of an episode, whereas delta-coverage is final coverage minus initial coverage.
All our coverage statistics are normalized (i.e., divided by the maximum possible coverage of the cloth in a flattened configuration) and can be easily computed from a top-down camera, while the number of interactions is the episode length.
To evaluate our policy, we load a task from the testing task datasets then run the policy for 10 steps or until the policy predicts grasps on the floor.

\begin{figure*}[t]
  \centering
  \includegraphics[width=\textwidth]{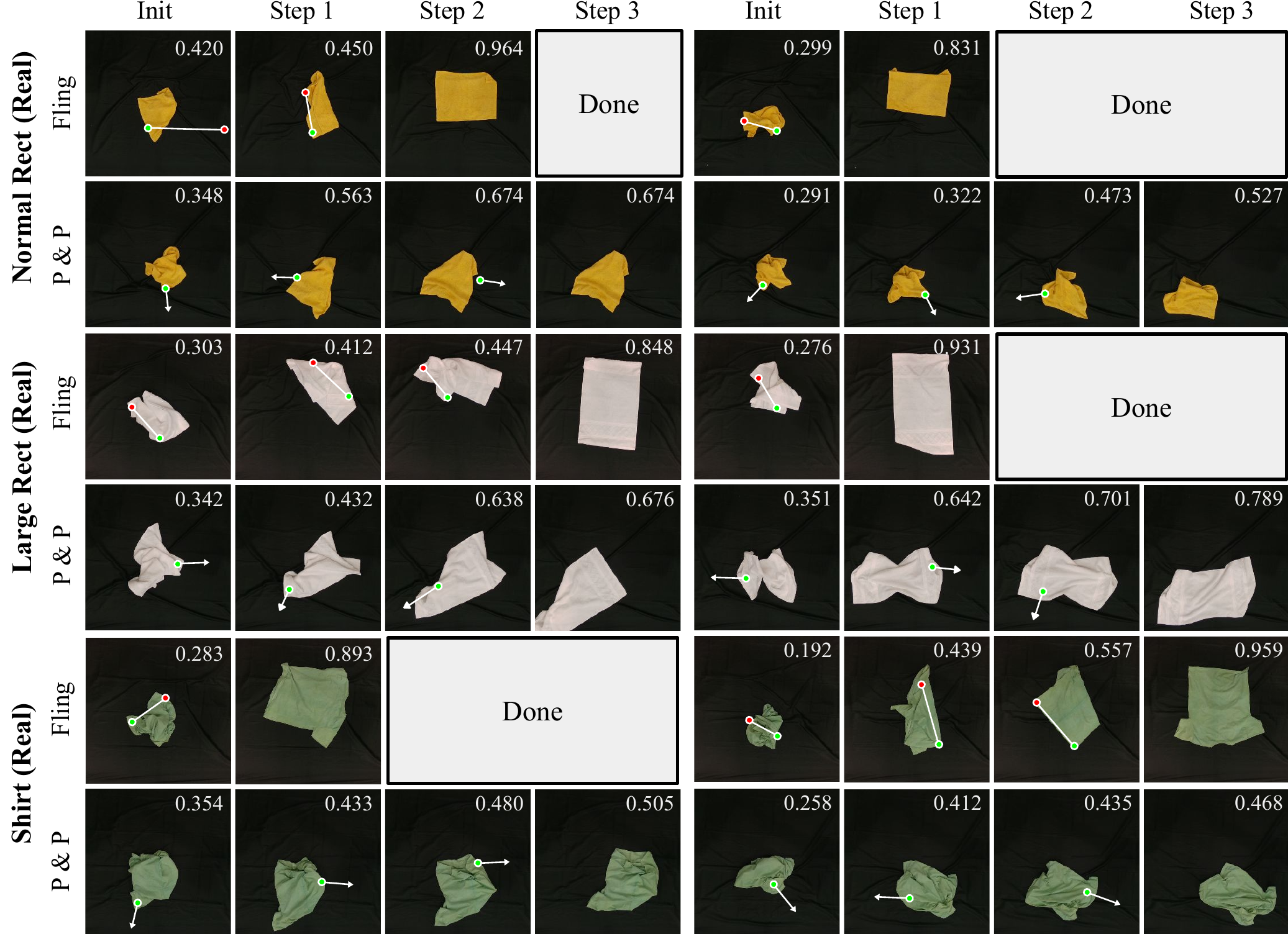}
  \caption{
    \textbf{Qualitative Results in Real World.}
    Cloth coverages are labeled on the top right corner. Red and green circles represent grasps by left and right arms, placed above and below field of view, respectively.
    \OURS discovered through trial-and-error a two-arm corner and edge grasp when corners and edges are visible.
    While the pick \& place baseline discovered a similar strategy, its performance is inherently limited by quasi-static actions, requiring significantly more steps to achieve a final coverage lower than \OURS's.
  }
  \label{fig:qualitative_results}
  \vspace{-6mm}
\end{figure*}

\vspace{-2mm}
\subsection{Task Dataset Generation}\label{sec:task_generation}
\vspace{-2mm}
Each task is specified by a cloth mesh, mass, stiffness, and initial configuration.
The cloth mesh is sampled from one of three types:
\vspace{-2mm}
\begin{enumerate}
  \item \textbf{Normal Rect}, which contains rectangular cloths with size within the reach range. Edge lengths are sampled from $[0.40\si{\m},0.65\si{\m}]$.
  \item \textbf{Large Rect}, which contains rectangular cloths with at least one edge larger than the reach range ($0.70\si{\m}$).
        Otherwise, edge sizes are sampled from $[0.40\si{\m},0.75\si{\m}]$, which means the shorter edge could still be smaller than the reach range.
  \item{
        \textbf{Shirt}, which contains a subset of shirts sampled from CLOTH3D's~\cite{bertiche2020cloth3d} test split, all of which are resized to be within the reach range.
        These shirts include tank tops, crop tops, short and long sleeves.
        } \vspace{-2mm}
\end{enumerate}

The cloth mass is sampled from $[0.2 \si{\kilogram},2.0\si{\kilogram}]$ and an internal stiffness from $[0.85 \si{\kilogram\per{\second^2}},0.95 \si{\kilogram\per{\second^2}}]$.
Finally, the cloth's initial configuration is varied by holding a randomly grasped the cloth at a random height between  $[0.5 \si{\meter},1.5\si{\meter}]$ then dropping and allowing the cloth to settle (similar to \citeauthor{lee2020learning} and tier-3 in \citeauthor{seita2019imitation}), resulting in a severely crumpled configuration.

To calculate the normalized coverage, we use the maximum possible coverage of the cloths in their flattened configurations.
For rectangular cloths in simulation, the flattened configuration can be analytically calculated using the undeformed vertex positions of the grid mesh, which means the normalized coverage could still be higher if the cloth rests in a stretched position due to friction.
For shirts in simulation, we opted to calculate its maximum possible coverage as its outer surface area divided by 2, since a qualitatively flattened T-shirt may not actually maximize coverage.
While this choice also makes it possible for the normalized coverage to be greater than 1, it will still preserve performance rankings.
For real world experiments, the flattened configurations are manually set.

We emphasize the difficulty of our unfolding tasks, where cloths in Normal Rect, Large Rect, and Shirt have average initial coverages of 28.8\%, 27.1\%, and 46.4\% in simulation respectively, and 26.1\%, 33.6\%, and 24.0\% in the real world respectively.
In contrast, simulated cloths from \citeauthor{seita2019imitation} start at $77.2\%$, $57.6\%$, and $42.0\%$ for easy, medium, and hard tasks, while real world cloths from \citeauthor{ganapathi2020learning} starts at 71.4\% and 68.4\% on two different real-world trials.
Yet, the challenging cloth configurations found in our tasks (Fig.~\ref{fig:qualitative_results}) are realistic and prevalent in typical households.

In simulation, the policy is trained on 2000 \emph{rectangular} cloths sampled evenly between Normal Rect and Large Rect, and evaluated on 600 novel tasks split evenly between Normal Rect, Large Rect, and Shirt cloths.
In real, the simulation policy is deployed to collect real world experience on 150 Normal Rect episodes (257 steps), optimized on both simulation and real world data, then evaluated on 10 novel test tasks in each cloth type.

\vspace{-2mm}
\subsection{Approach Comparisons}\label{sec:comparisons}
\vspace{-2mm}
\begin{itemize}
  \item \textbf{Quasi-static}:
        [Pick \& Place] and [Pick \& Drag] are policies which uses a quasi-static ``pick and place'' (similar to \citet{lee2020learning}, visualized in Fig.~\ref{fig:primitives}b) and ``pick and drag'' (similar to \citeauthor{seita_bags_2021}, with no lift step compared to pick and place) primitive respectively.
          [Stretch \& Drag] is identical to [Pick \& Drag] with an extra stretch step (identical to FlingBot's stretch) after picking.
        These baselines are implemented with dual-arm set up, therefore provides the same physical reach range as our dual-arm fling system.
  \item \textbf{Dynamic manipulation}:
        [FlingBot] is our policy which predicts the optimal two-arm grasp locations for a fixed stretching and flinging routine described in Sec.~\ref{sec:motion_primitives}.

  \item \textbf{Dynamic manipulation with fling speed prediction}:
        [FlingBot-S] is identical to [FlingBot] with an additional fling speed module, which predicts the fling speed within the range of $[0.1\si{\meter\per\second},1.0\si{\meter\per\second}]$ from visual input of the cloth after stretching and lifting.
        This fling speed module is trained using Deep Deterministic Policy Gradients (DDPG)~\cite{lillicrap2015continuous} on the delta-coverage rewards to maximize single-step returns (discount factor $\gamma$ is set to 0), where the grasps are predicted by a converged and frozen [FlingBot] value network.
        However, despite the cloth physical parameter variations in the training and testing dataset (Sec.~\ref{sec:task_generation}), Tab.~\ref{tab:sim} and Fig.~\ref{fig:episode_length} suggest that there aren't significant performance gains for [FlingBot-S] over [FlingBot] within the cloth parameter ranges tested (Sec.~\ref{sec:task_generation}).
        These results justify [FlingBot]'s usage of a single fling speed, even for varying cloth mass, stiffness, and size.
        Thus, we prefer the simpler [FlingBot] approach for comparisons with baselines.
  \item \textbf{Dynamic manipulation with fling parameter regression}:
        [Fling-Reg] is identical to [FlingBot], but directly predicts the action parameters $\langle C_x,C_y,\theta,w \rangle$ from visual inputs instead of exploiting the task's equivariances.
        Its policy is trained using DDPG on delta-coverage rewards to maximize single-step returns.
        However, from Tab.~\ref{tab:sim}, [Fling-Reg] completely fails to perform the task, demonstrating the advantage of encoding inductive biases which leverage equivariances in the problem structure.
\end{itemize}

\vspace{-4mm}
\subsection{Results}\label{sec:results}
\vspace{-2mm}

\begin{figure*}[t]
  \centering
  \begin{subfigure}[b]{\textwidth}
    \center
    \includegraphics[width=0.95\textwidth,trim=0 10 0 0, clip]{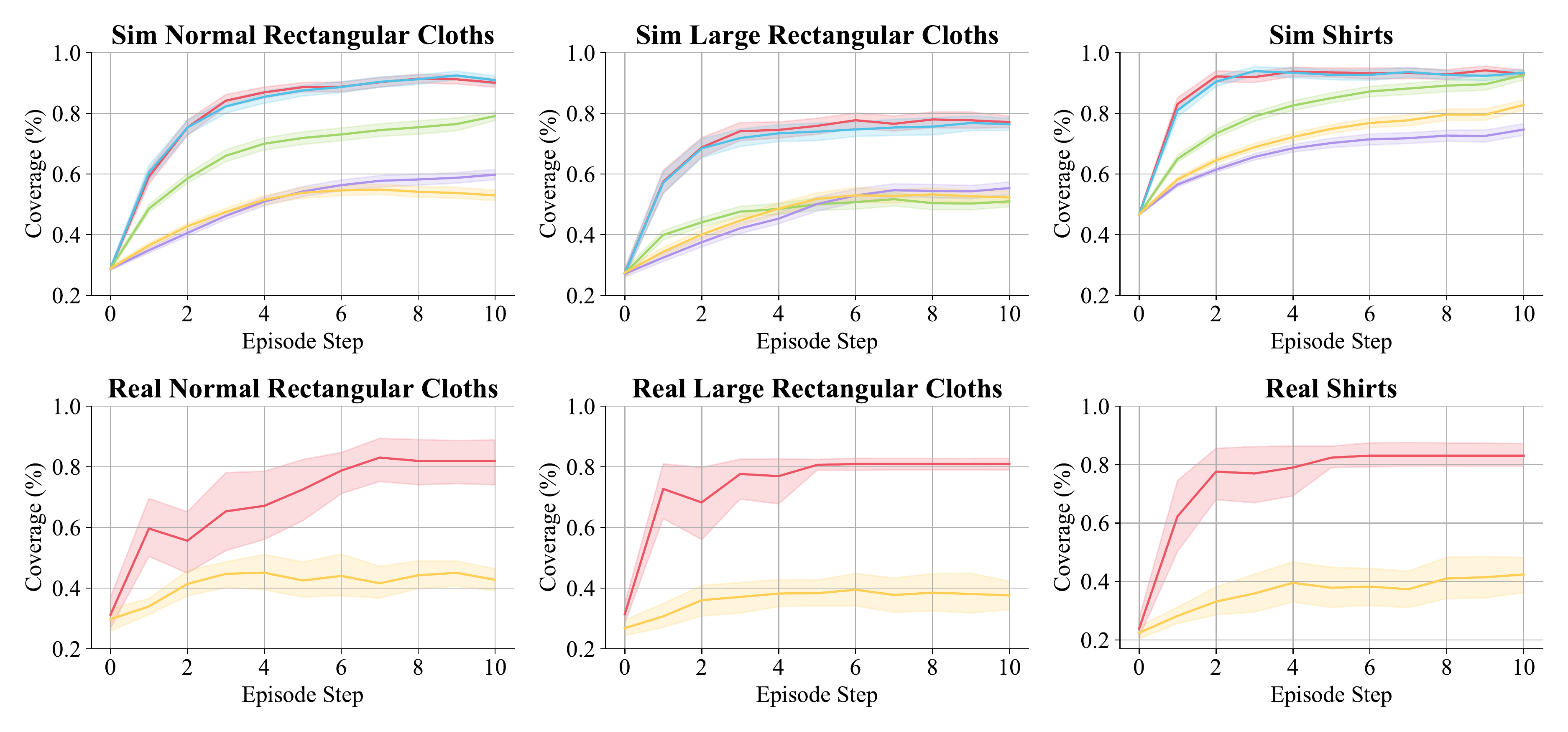}
  \end{subfigure}
  \begin{subfigure}[b]{\textwidth}
    \center
    \vspace{-15mm}
    \includegraphics[width=0.95\textwidth,trim=90 10 90 465, clip]{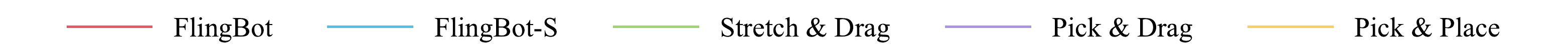}
  \end{subfigure}
  \vspace{-4mm}
  \caption{\textbf{Coverage v.s. Steps.} With 95\% confidence interval shaded. FlingBot can achieve high coverage within a few interaction steps, while the quasi-static baselines never reach high coverages even with significantly more interaction steps. This demonstrates the difficulty of unfolding from highly crumpled initial configurations and dynamic action's superior efficiency.}
  \label{fig:episode_length}
  \vspace{-5mm}
\end{figure*}

\mypara{Better Efficiency.}
In this experiment, we compare the unfolding efficiency between quasi-static and dynamic actions.
From the Normal Rect column in Tab.~\ref{tab:sim},  [FlingBot] increases the coverage of the cloth ($+63.1\%$) by more than two times that of [Pick\&Place] and [Pick\&Drag] baselines ($+29.2 \%$, and $+24.2 \%$).
Additionally, from Fig.~\ref{fig:episode_length}, [FlingBot] achieves over $80\%$ within 3 interactions (simulation normal cloth), while the quasi-static baselines never reach such a high coverage even with significantly more interaction steps or with a stretching subroutine.
This demonstrates the difficulty of unfolding from highly crumpled initial configurations and dynamic action's superior efficiency.

\begin{wraptable}{r}{6.5cm}
  \small
  \setlength{\tabcolsep}{0.04cm}
  \vspace{-4mm}
  \begin{tabular}{lccccccc}
    \toprule
                  & Normal Rect                   & Large Rect                    & Shirt                         \\\midrule
    Pick\&Place   & 53.0 / 24.2                   & 52.0 / 24.8                   & 79.8 / 33.4                   \\
    Pick\&Drag    & 58.0 / 29.2                   & 54.2 / 27.1                   & 72.4 / 26.0                   \\
    Stretch\&Drag & 77.2 / 48.4                   & 50.2 / 23.1                   & 90.3 / 43.9                   \\
    Fling-Reg     & 29.1 / 0.3                    & 27.7 / 0.5                    & 54.0 / 7.6                    \\
    \midrule
    FlingBot-S    & \textbf{92.6} / \textbf{63.8} & 78.9 / 51.7                   & \textbf{93.5} / \textbf{47.1} \\
    \OURS         & 91.9 / 63.1                   & \textbf{79.2} / \textbf{52.0} & 93.3 / 46.8                   \\
    \bottomrule
  \end{tabular}
  \captionsetup{justification=centering}
  \caption{
  Simulation Experiments \\
    (Final / Delta Coverage). 
    }
  \label{tab:sim}
  \vspace{-6mm}
\end{wraptable}

\mypara{Increased Reach Range.}
In this experiment, we investigate these approaches' performance on cloths which have dimensions larger than the robot arm's reach range. These tasks are not only challenging for quasi-static baselines, because the arms can't manipulate the cloth at locations beyond its reach range, but also for our flinging policies, because the arms can't fully stretch or lift the cloth off the ground.
Despite these challenges, [FlingBot] achieves $79.2\%$ (Tab.~\ref{tab:sim}, column Large Rect).
Compared to the quasi-static baselines, [FlingBot] increases the coverage by $+52.0\%$, which is roughly twice that of the quasi-static baselines ( $+27.1\%$, $+24.8\%$, $+23.1\%$).
These results show how high-velocity actions could effectively expand the physical reach range, allowing the system to be more flexible with extreme cloth sizes.

\mypara{Generalize to Unseen Cloth Types.}
In this experiment, we investigate how well these approaches, trained on only rectangular cloths, can generalize to unseen cloth types (i.e.: T-shirts).
Qualitative real world (Fig.~\ref{fig:qualitative_results}) and simulation
\makeatletter
\@ifundefined{r@fig:sim_qualitative_results}{%
    (in supplementary materials)
  }{%
    (Fig.~\ref{fig:sim_qualitative_results})
  }%
\makeatother
results suggest that our flinging policy has learned to grasp keypoints on cloth (i.e., corners, edges) when it sees them, or otherwise fling to reveal these features.
\OURS's generalization performance (93.3\% in Tab.~\ref{tab:sim}, column Shirt) to novel cloth geometries can be attributed to this strategy, since a cloth of any type can be unfolded by grabbing one of its edges, stretching, then flinging it in a direction perpendicular to the edge.
Through self-supervised exploration, our approach discovered grasping strategies which were manually designed in heuristic based prior works, while being more generalizable to different cloth configurations and types.
Meanwhile, all quasi-static baselines exhibited worse cloth unfolding efficiency.
The Sim Shirts plot in Fig.~\ref{fig:episode_length} shows that our flinging policies take only 3 actions to reach their maximum coverages, while the quasi-static baselines take upwards of 8 steps to reach lower maximum coverages.

\begin{wraptable}[7]{r}{6cm}
  \vspace{-2mm}
  \small
  \tabcolsep=0.04cm
  \centering
  \vspace{-5mm}
  \begin{tabular}{lccccccc}                                                                                                  \\\toprule
    Action      & {Normal Rect}                 & {Large Rect}                  & {Shirt}                       \\
    \midrule
    Pick\&Place & 43.2 / 13.0                   & 38.4 / 11.0                   & 42.7 / 19.9                   \\
    \OURS       & \textbf{81.9} / \textbf{55.8} & \textbf{88.5} / \textbf{54.9} & \textbf{89.2} / \textbf{65.2} \\
    \bottomrule
  \end{tabular}
  \captionsetup{justification=centering}
  \caption{
    Real World Experiment \\
    (Final / Delta Coverage).
  }\label{tab:real}
  \vspace{-5mm}
\end{wraptable}

\mypara{Evaluating Real-World Unfolding.}
Finally, we finetune and evaluate our simulation models from Tab.~\ref{tab:sim} with real-world experience on a pair of UR5 arms.
Task generation is automated using the robot arms by randomly grasping the cloth at height $0.50\si{\meter}$ then dropping it back on the workspace.
We use a $0.35\si{\meter}\times0.45\si{\meter}$ cloth for Normal Rect, a $0.40\si{\meter}\times0.70\si{\meter}$ bath towel for Large Rect, and a $0.45\si{\meter}\times0.55\si{\meter}$ T-shirt for Shirt.
The system collected 257 experience steps over 150 cloth tasks for finetuning in total.
The performance is reported averaged over 10 test episodes, where real-world grasp errors are filtered out (see ``Real World Failure Cases'' below).
From Tab.~\ref{tab:real}, we report that our policy achieves over $80\%$ on all cloth types, which outperforms the quasi-static pick \& place baseline by over $40\%$.
Additionally, we report that the pre-finetune performance of FlingBot on Normal Cloths is $69.8\%$, justifying our decision to finetune to get a $12.1\%$ improvement.
Overall, our flinging primitive takes a median time of $16.7\si{\second}$.
While the pick \& place primitive only takes a median time of $8.8\si{\second}$, it is unable to reach the high coverages even with many more interaction steps (Fig.~\ref{fig:episode_length}, bottom row).
Both primitives incur additional overheads of $2.5\si{\second}$ for preparing the transformed observation batch and $2.9\si{\second}$ for a routine which checks whether the cloth is stuck to the gripper after the gripper is opened (details in 
\makeatletter\@ifundefined{r@sec:real_world_failures}{supplementary materials}{Sec.~\ref{sec:real_world_failures}}\makeatother).

\mypara{Real World Failure Cases.}\label{para:failure}
Grasping failures, where the policy specified a grasp point on the cloth but the grippers failed to successful pinch grasp, constituted all of our real-world pipeline failure cases. 
Our real world system uses its frontal RGBD view to detect grasp failures after the dual arm lifts the cloths up and automatically discards these episodes.
The average grasp success rate is 78.0\%, 45.0\%, and 75.8\% for normal rectangular, large rectangular, and shirts respectively.
We used a bath towel for our large rectangular cloths, which is significantly thicker and stiffer than the tea towel and T-shirt we used for normal rect and shirts. Therefore, pinch grasps with large cloths failed significantly more.
Additionally, even on successful grasps, the gripper may hold the cloth tight in its crumpled state, rendering all flings ineffective.
However, we do not discard of these episodes.
\makeatletter\@ifundefined{r@sec:real_world_failures}{
We discuss more of real world grasp failures and show simulation failure cases in the supplementary materials
}{
We discuss more of real world grasp failures in Sec.~\ref{sec:real_world_failures} and show simulation failure cases in Fig.~\ref{fig:failure_cases}}\makeatother.

%% file: text/conclusion.tex
\section{Conclusion and Future Work}
\vspace{-3mm}
We proposed a dynamic fling motion primitive and a self-supervised learning algorithm for learning the grasp parameters for the cloth unfolding task.
The cloth unfolding policy is efficient, generalizable, and works with cloth sizes beyond the reach range of the system in both simulation and the real world.
While we've demonstrated the effectiveness of dynamic actions for cloth unfolding, dynamic actions alone are insufficient for more complex deformable object manipulation tasks, such as goal-conditioned folding.
In this direction, future work could explore learning in combined action spaces (i.e., dynamic and quasi-static, two-arm and single-arm, etc.)  and integrating it with vision-based cloth pose estimation~\cite{chi2021garmentnets} for goal-conditioned cloth manipulation.

%% file: text/supp.tex
\section{Supplementary Materials}

\subsection{Extra Qualitative Results}

\begin{figure*}[h]
    \centering
    \includegraphics[width=\textwidth]{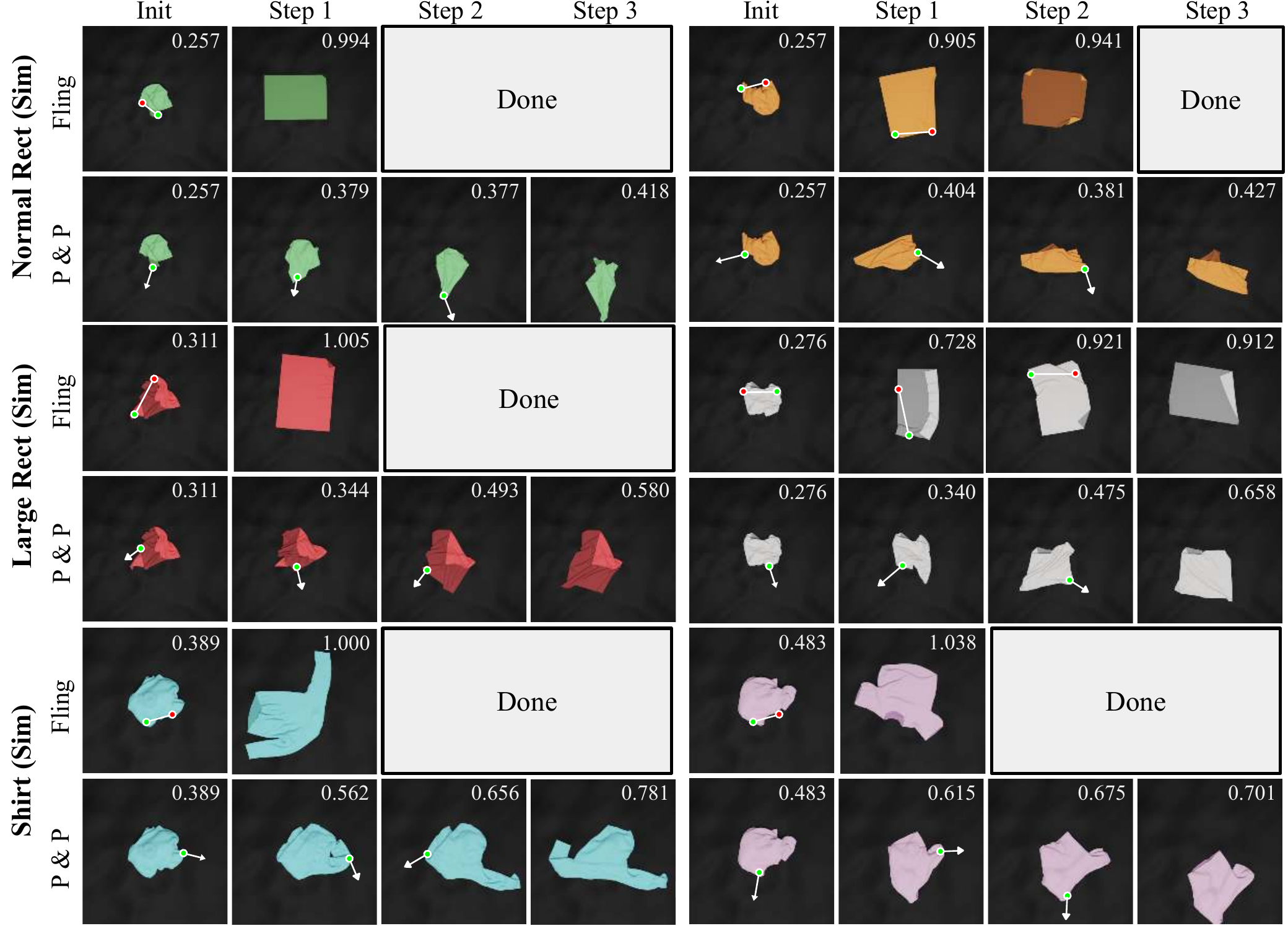}
    \caption{
        \textbf{\textbf{Qualitative Results in Simulation Experiments.} }
    }
    \label{fig:sim_qualitative_results}
    \vspace{-5mm}
\end{figure*}

\begin{wrapfigure}{r}{0.45\textwidth}
    \vspace{-6mm}
    \centering
    \includegraphics[width=\linewidth]{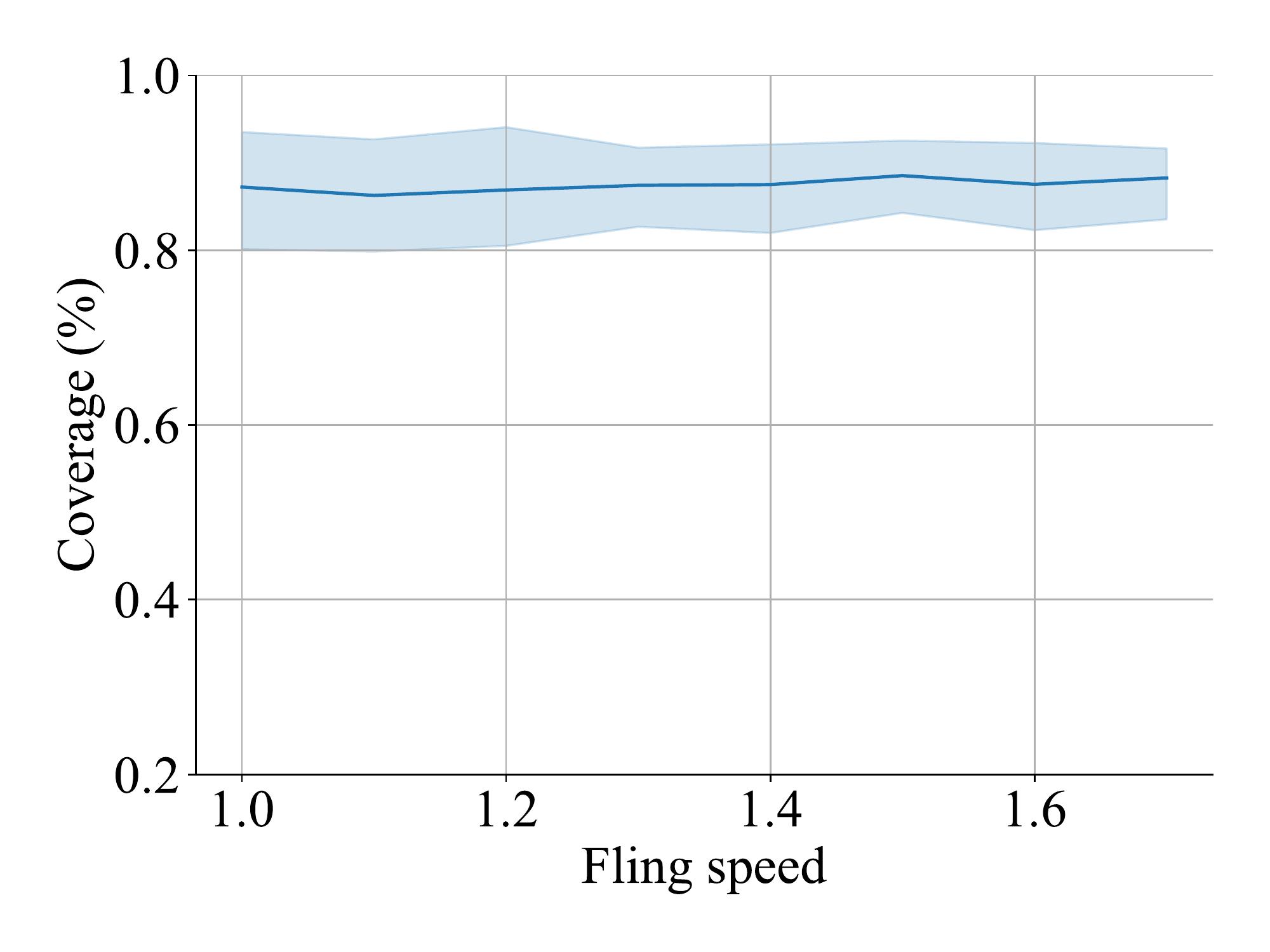}
    \vspace{-6mm}
    \caption{
        \textbf{Real world fling speed robustness.}
        By flinging at speeds in the range $[1.0 \si{\meter\per\second},1.7\si{\meter\per\second}]$ at $0.1\si{\meter\per\second}$ intervals, we observed that our fling primitive robustly achieves above $80\%$ coverage if given two good grasps.
    }
    \vspace{-2mm}
    \label{fig:real_world_fling_robustness}
\end{wrapfigure}

\subsection{Failure cases}

For normal rectangular cloths, the most common failure case is when a dual-arm corner grasped is slightly misaligned and becomes a single-arm grasp and fling instead, resulting in a low coverage configuration.
For large rectangular cloths, cloths could fold in half almost perfectly, thus appearing completely unfolded, causing the policy to terminate the episode.
For shirts, the self-discovered dual-arm corner and edge grasp for flinging which is effective for rectangular cloths fail on shirts in two main ways.
First, if the sleeves of the shirt get stuck in the shirt’s collar, FlingBot will be unable to pull the sleeves out.
This failure case motivates future work on combining quasi-static and dynamic actions for cloth manipulation.
Second, dual arm grasps where one grasp is on the outer surface and another grasp is on the inner surface of the shirt usually flings to low coverages.
While this failure case is expected to the differences between rectangular cloths and shirts (presence of holes, inner/outer surfaces, etc.), FlingBot’s performance on shirts still suggested generalizable cloth manipulation abilities.

\begin{figure*}[h]
    \centering
    \includegraphics[width=\textwidth, trim=80 0 0 0, clip]{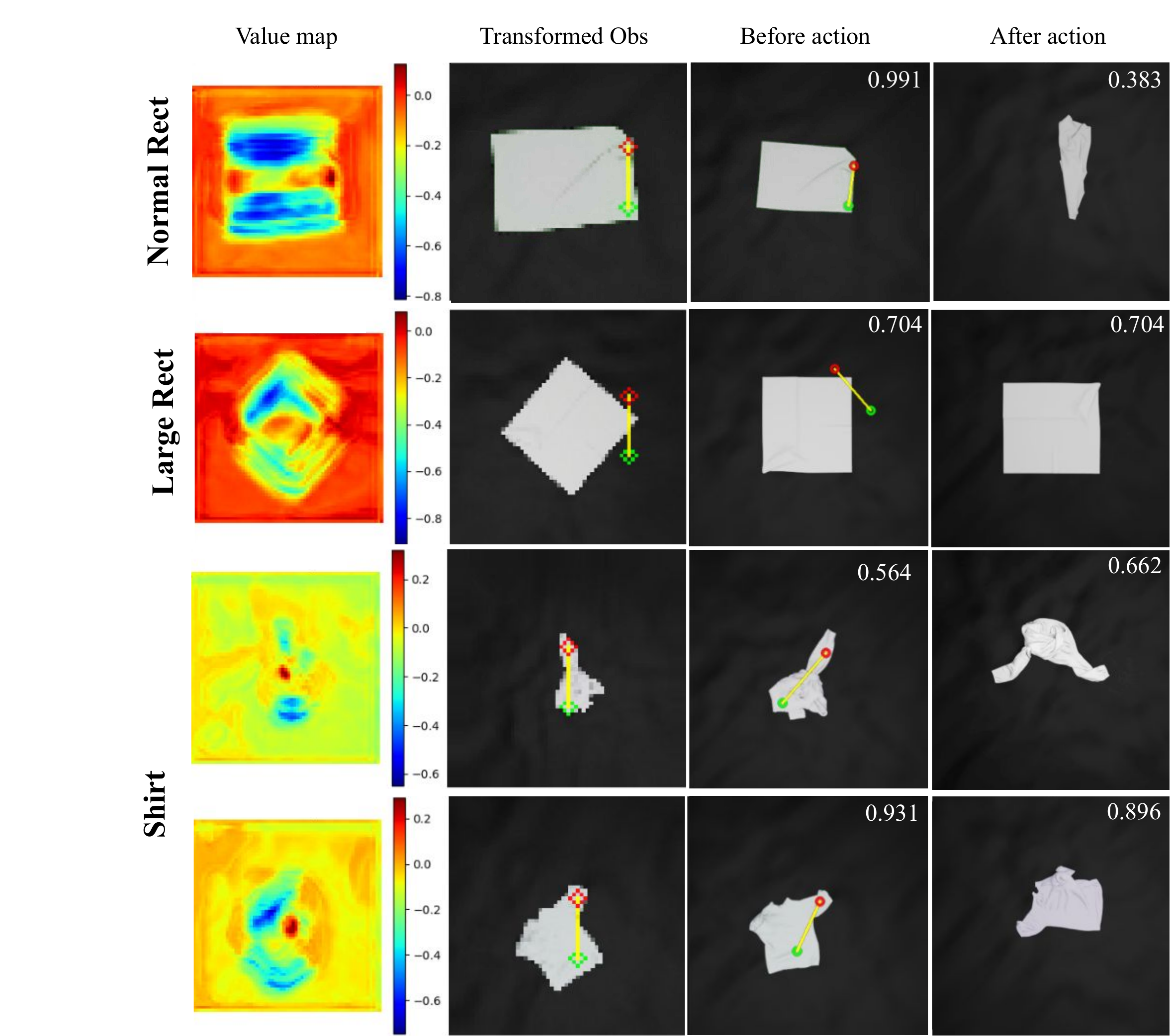}
    \caption{
        \textbf{\textbf{Failure Cases in Simulation Experiments.} }
    }
    \label{fig:failure_cases}
    \vspace{-5mm}
\end{figure*}

\subsection{Real world fling parameter robustness}
\label{sec:real_world_fling_speed}

In designing our motion primitive, we optimized fling parameters (waypoints, velocities, acceleration) to maximize coverage assuming a good grasp (e.g.: a dual arm grasp on a normal rectangular cloth in a stretched state).
We observed that the real world flinging setup system could robustly unfold the cloth using a wide range of fling parameters (i.e.: fling heights and speeds)  if manually given a good grasp (Fig~\ref{fig:real_world_fling_robustness}), while the simulated system was highly sensitive to such parameters. 
This sim2real gap was bridged in our work by tuning the simulated fling parameters such that flings from good grasps in simulation would also lead to high coverages like in the real system, which results in a variable fling height and a different fling speed in simulation compared to a fixed fling height and speed in real.
Crucially, this gap underscores the importance of real world results for cloth manipulation as well as motivates future work on fast and accurate cloth simulation engines.

Another large sim2real problem was poor collision handling in simulation. There were cloth configurations that the real world system experienced but was not observed in simulation, such as cloth twisting. While Nvidia Flex is decent at preventing self penetration, it does so at the cost of unrealistic collision handling. Qualitatively, this unrealistic collision handling means cloths untwist themselves when twisted (or are in any other state with high self-collision). Another case where this self-unfolding behavior was observed was for shirts due to the two layers of the shirts colliding with each other. We hypothesize that poor collision handling is the main reason why performance for the simulated shirt benchmark is higher across the board for all approaches, despite being unseen cloth types.

\subsection{Real world Failures}
\label{sec:real_world_failures}

All failure cases in our real world experiment were due to failed grasps (where the policy specified a grasp point on the cloth but the grippers failed to pinch grasp the cloth). Cloth grasping failure is a common problem when working with real world cloth manipulation. For instance, \citeauthor{ganapathi2020learning} also observed that “the most frequent failure mode is an unsuccessful grasp of the fabric which is compounded for tasks that require more actions”. However, we believe this issue can be mitigated by using specialized gripper hardware (like in \citeauthor{ganapathi2020learning,seita2019imitation}) or incorporating grasp success estimation.

In addition to grasping failures discussed in \makeatletter\@ifundefined{r@para:failure}{the main paper}{Sec~\ref{para:failure}}\makeatother, we also observed occasional cloth stuck errors, where cloths get stuck to the gripper after the gripper opens in an attempt to release the cloth.
To enable the system to automatically recover from this issue, after opening the gripper, the arms move to a predefined height above the workspace.
Then, we use the frontal RGB-D view (used to implement the stretching primitive) to check whether the cloth is detected above a manually set height threshold above the workspace.

Here, we summarize techniques for implementing a real-world cloth manipulation pipeline:
\begin{enumerate}
    \item \textbf{High friction gripper finger}:
          In our dual-arm system, we used an OnRobot RG2 and Schunk WGS50 gripper, where the former had a rubber tip, and the latter had a metal tip.
          After observing a significantly lower pinch grasping success with the WSG50, we added a rubber fingertip to the WSG50, which improved its grasp successes significantly.
          Alternatively, picking a gripper that can apply lots of pressure using its fingers, like the da Vinci Research Kit surgical robot in \citeauthor{seita2019imitation}, should also help.
    \item \textbf{Soft Workspace}: A successful pinch grasp should apply the right amount of pressure between the cloth and the workspace.
          Too little pressure and the cloth will not get grasped, while too much pressure could easily damage the cloth, gripper, and robot arm.
          Due to noise in-depth sensing, the arms may be asked to grasp points slightly below the surface of the workspace.
          Similar to prior works \cite{ganapathi2020learning, seita2019imitation}, we found that using a firm and thin rubber mattress was sufficient to address this problem.
 \item \textbf{Accurate depth sensing}: Building on the previous point, hardware improvements on the sensing side could help significantly.
 In our real world experiments, the Azure Kinect v3 has significantly less noisy and more accurate depth images than the Intel Realsense D415.
 While the real world numbers we reported in this paper are only with Realsense cameras, the codebase for our Kinect/Realsense real world setup is publicly accessible at \href{https://github.com/columbia-ai-robotics/flingbot}{https://github.com/columbia-ai-robotics/flingbot}.
    \item \textbf{4 DOF Grasp success prediction}:
          A grasp which is parallel to a small crease on the cloth is more likely to result in a successful pinch grasp.
          Therefore, we hypothesize that additionally considering gripper z-rotation, on top of the positional 3 DOFs we have in our system, and learning the optimal pinch grasp z-rotation using a grasp success predictor may result in higher overall grasp success.
          Future work could explore weighing task rewards with such grasp success predictions.
\end{enumerate}

\begin{figure}
    \centering
    \begin{subfigure}[b]{0.45\textwidth}
        \centering
        \includegraphics[width=\textwidth, trim=70 200 1000 200, clip]{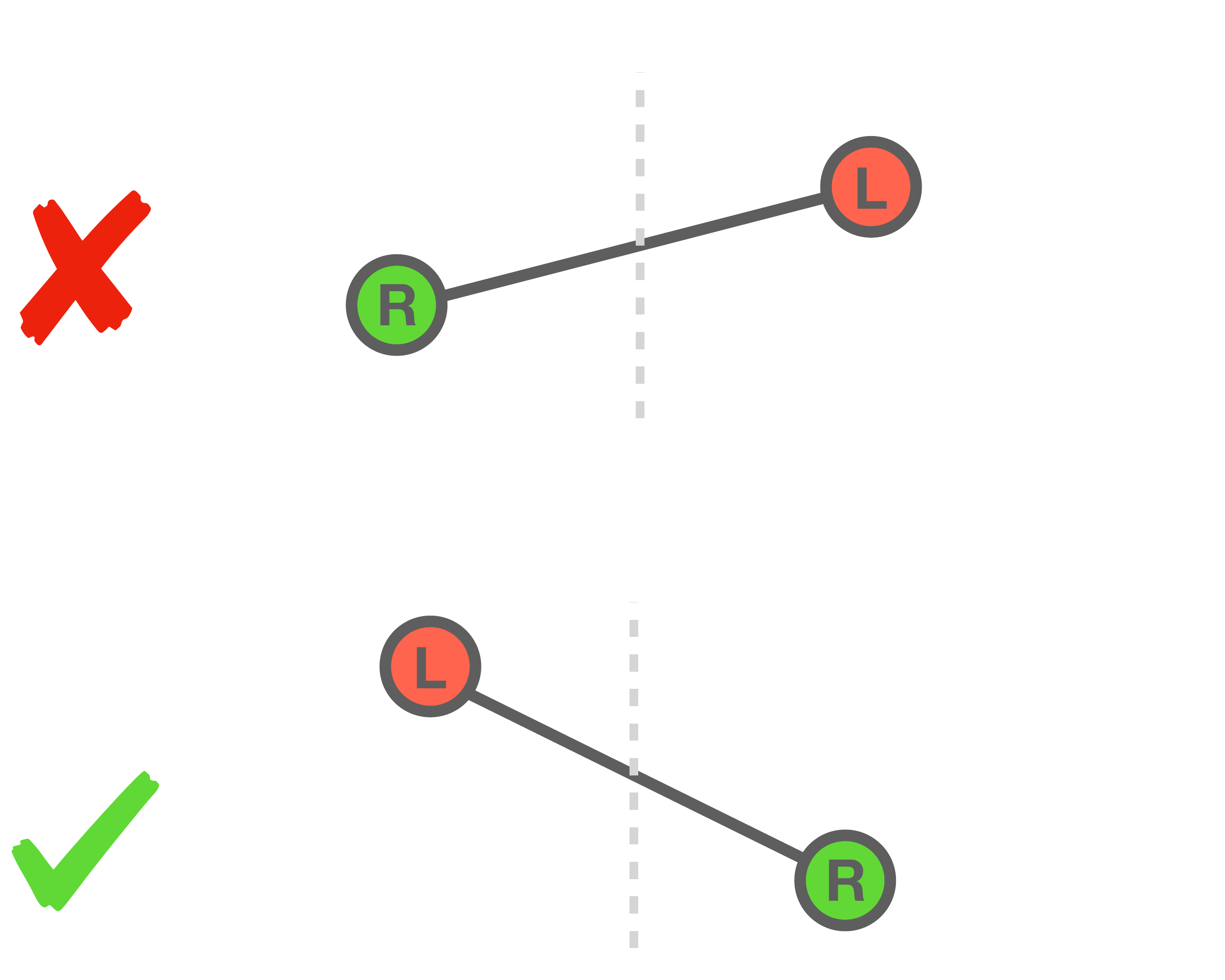}
        \caption{Crossover constraint}
        \label{fig:crossover_constraint}
    \end{subfigure}
    \hfill
    \begin{subfigure}[b]{0.45\textwidth}
        \centering
        \includegraphics[width=\textwidth, trim=900 200 70 200, clip]{figures/dual-arm-constraints.pdf}
        \caption{Grasp width constraint}
        \label{fig:grasp_width_constraint}
    \end{subfigure}
    \caption{
        To minimize collisions, arms should grasp points closer to their side (a) and be a reasonable distance away from each other (b).
    }
    \label{fig:dual_arm_constraints}
\end{figure}

\subsection{Designing dynamic motion primitives}

To achieve the highest speed at the end effector while respecting the torque limit of each joint, the primitive must move upper joints (i.e.: wrist) more than the lower joints (i.e., base). We also found that adding a blending radius between each target joint configuration gave a much smoother flinging trajectory and cloth swinging, as opposed to a jerky cloth motion without blending radius. In designing our motion primitive, we optimized fling dynamics parameters (waypoints, velocities, acceleration) to maximize coverage assuming a dual arm grasp on a normal rectangular cloth in a stretched state. Automatically discovering dynamic motion primitives, such as flinging, and simultaneously learning their parameters is an important and interesting direction for future work.

\subsection{Why is learning fling speeds unhelpful?}

We hypothesize that light and thin cloths, whose air resistive forces to momentum during flinging ratio is significantly higher than the cloths tested, would require a higher fling speed. Therefore, fling speed learning may be helpful when cloths in the task dataset contain a wider variance in density and thickness. We also hypothesize that a successful fling speed prediction approach may require extra information about the cloths’ physical parameters that could not be obtained through visual observation alone, which would be out of scope for this work.

%% file: paper.bbl
\begin{thebibliography}{25}
\providecommand{\natexlab}[1]{#1}
\providecommand{\url}[1]{\texttt{#1}}
\expandafter\ifx\csname urlstyle\endcsname\relax
  \providecommand{\doi}[1]{doi: #1}\else
  \providecommand{\doi}{doi: \begingroup \urlstyle{rm}\Url}\fi

\bibitem[{Mason} and {Lynch}(1993)]{dynamic1993mason}
M.~T. {Mason} and K.~M. {Lynch}.
\newblock Dynamic manipulation.
\newblock In \emph{Proceedings of 1993 IEEE/RSJ International Conference on
  Intelligent Robots and Systems (IROS '93)}, volume~1, pages 152--159 vol.1,
  1993.
\newblock \doi{10.1109/IROS.1993.583093}.

\bibitem[Lee et~al.(2020)Lee, Ward, Cosgun, Dasagi, Corke, and
  Leitner]{lee2020learning}
R.~Lee, D.~Ward, A.~Cosgun, V.~Dasagi, P.~Corke, and J.~Leitner.
\newblock Learning arbitrary-goal fabric folding with one hour of real robot
  experience, 2020.

\bibitem[Seita et~al.(2019)Seita, Ganapathi, Hoque, Hwang, Cen, Tanwani,
  Balakrishna, Thananjeyan, Ichnowski, Jamali, Yamane, Iba, Canny, and
  Goldberg]{seita2019imitation}
D.~Seita, A.~Ganapathi, R.~Hoque, M.~Hwang, E.~Cen, A.~K. Tanwani,
  A.~Balakrishna, B.~Thananjeyan, J.~Ichnowski, N.~Jamali, K.~Yamane, S.~Iba,
  J.~F. Canny, and K.~Goldberg.
\newblock Deep imitation learning of sequential fabric smoothing policies.
\newblock \emph{CoRR}, abs/1910.04854, 2019.
\newblock URL \url{http://arxiv.org/abs/1910.04854}.

\bibitem[Wu et~al.(2019)Wu, Yan, Kurutach, Pinto, and
  Abbeel]{wu2019withoutdemonstrations}
Y.~Wu, W.~Yan, T.~Kurutach, L.~Pinto, and P.~Abbeel.
\newblock Learning to manipulate deformable objects without demonstrations.
\newblock \emph{CoRR}, abs/1910.13439, 2019.
\newblock URL \url{http://arxiv.org/abs/1910.13439}.

\bibitem[Zeng et~al.(2018)Zeng, Song, Welker, Lee, Rodriguez, and
  Funkhouser]{zeng2018learning}
A.~Zeng, S.~Song, S.~Welker, J.~Lee, A.~Rodriguez, and T.~Funkhouser.
\newblock Learning synergies between pushing and grasping with self-supervised
  deep reinforcement learning.
\newblock 2018.

\bibitem[Zeng et~al.(2019)Zeng, Song, Lee, Rodriguez, and
  Funkhouser]{zeng2019tossingbot}
A.~Zeng, S.~Song, J.~Lee, A.~Rodriguez, and T.~Funkhouser.
\newblock Tossingbot: Learning to throw arbitrary objects with residual
  physics.
\newblock 2019.

\bibitem[Wu et~al.(2020)Wu, Sun, Zeng, Song, Lee, Rusinkiewicz, and
  Funkhouser]{Wu_2020}
J.~Wu, X.~Sun, A.~Zeng, S.~Song, J.~Lee, S.~Rusinkiewicz, and T.~Funkhouser.
\newblock Spatial action maps for mobile manipulation.
\newblock \emph{Robotics: Science and Systems XVI}, Jul 2020.
\newblock \doi{10.15607/rss.2020.xvi.035}.
\newblock URL \url{http://dx.doi.org/10.15607/RSS.2020.XVI.035}.

\bibitem[Li et~al.(2019)Li, Wu, Tedrake, Tenenbaum, and
  Torralba]{li2019learning}
Y.~Li, J.~Wu, R.~Tedrake, J.~B. Tenenbaum, and A.~Torralba.
\newblock Learning particle dynamics for manipulating rigid bodies, deformable
  objects, and fluids.
\newblock In \emph{ICLR}, 2019.

\bibitem[Lin et~al.(2020)Lin, Wang, Olkin, and Held]{lin2020softgym}
X.~Lin, Y.~Wang, J.~Olkin, and D.~Held.
\newblock Softgym: Benchmarking deep reinforcement learning for deformable
  object manipulation, 2020.

\bibitem[Sun et~al.(2013)Sun, Aragon-Camarasa, Cockshott, Rogers, and
  Siebert]{Sun2013AHA}
L.~Sun, G.~Aragon-Camarasa, W.~Cockshott, S.~Rogers, and J.~Siebert.
\newblock A heuristic-based approach for flattening wrinkled clothes.
\newblock In \emph{TAROS}, 2013.

\bibitem[{Willimon} et~al.(2011){Willimon}, {Birchfield}, and
  {Walker}]{willimon2011model}
B.~{Willimon}, S.~{Birchfield}, and I.~{Walker}.
\newblock Model for unfolding laundry using interactive perception.
\newblock In \emph{2011 IEEE/RSJ International Conference on Intelligent Robots
  and Systems}, pages 4871--4876, 2011.
\newblock \doi{10.1109/IROS.2011.6095066}.

\bibitem[Seita et~al.(2018)Seita, Jamali, Laskey, Berenstein, Tanwani,
  Baskaran, Iba, Canny, and Goldberg]{seita2018bedmaking}
D.~Seita, N.~Jamali, M.~Laskey, R.~Berenstein, A.~K. Tanwani, P.~Baskaran,
  S.~Iba, J.~F. Canny, and K.~Goldberg.
\newblock Robot bed-making: Deep transfer learning using depth sensing of
  deformable fabric.
\newblock \emph{CoRR}, abs/1809.09810, 2018.
\newblock URL \url{http://arxiv.org/abs/1809.09810}.

\bibitem[Maitin-Shepard et~al.(2010)Maitin-Shepard, Cusumano-Towner, Lei, and
  Abbeel]{maitin2010cloth}
J.~Maitin-Shepard, M.~Cusumano-Towner, J.~Lei, and P.~Abbeel.
\newblock Cloth grasp point detection based on multiple-view geometric cues
  with application to robotic towel folding.
\newblock In \emph{2010 IEEE International Conference on Robotics and
  Automation}, pages 2308--2315. IEEE, 2010.

\bibitem[Triantafyllou et~al.(2016)Triantafyllou, Mariolis, Kargakos,
  Malassiotis, and Aspragathos]{triantafyllou2016geometric}
D.~Triantafyllou, I.~Mariolis, A.~Kargakos, S.~Malassiotis, and N.~Aspragathos.
\newblock A geometric approach to robotic unfolding of garments.
\newblock \emph{Robotics and Autonomous Systems}, 75:\penalty0 233--243, 2016.

\bibitem[Yuba et~al.(2017)Yuba, Arnold, and Yamazaki]{yuba2017unfolding}
H.~Yuba, S.~Arnold, and K.~Yamazaki.
\newblock Unfolding of a rectangular cloth from unarranged starting shapes by a
  dual-armed robot with a mechanism for managing recognition error and
  uncertainty.
\newblock \emph{Advanced Robotics}, 31\penalty0 (10):\penalty0 544--556, 2017.

\bibitem[Ganapathi et~al.(2020)Ganapathi, Sundaresan, Thananjeyan, Balakrishna,
  Seita, Grannen, Hwang, Hoque, Gonzalez, Jamali, Yamane, Iba, and
  Goldberg]{ganapathi2020learning}
A.~Ganapathi, P.~Sundaresan, B.~Thananjeyan, A.~Balakrishna, D.~Seita,
  J.~Grannen, M.~Hwang, R.~Hoque, J.~E. Gonzalez, N.~Jamali, K.~Yamane, S.~Iba,
  and K.~Goldberg.
\newblock Learning dense visual correspondences in simulation to smooth and
  fold real fabrics, 2020.

\bibitem[Jangir et~al.(2019)Jangir, Alenya, and Torras]{jangir2019dynamic}
R.~Jangir, G.~Alenya, and C.~Torras.
\newblock Dynamic cloth manipulation with deep reinforcement learning.
\newblock \emph{arXiv preprint arXiv:1910.14475}, 2019.

\bibitem[Balaguer and Carpin(2011)]{balaguer2011combining}
B.~Balaguer and S.~Carpin.
\newblock Combining imitation and reinforcement learning to fold deformable
  planar objects.
\newblock In \emph{2011 IEEE/RSJ International Conference on Intelligent Robots
  and Systems}, pages 1405--1412. IEEE, 2011.

\bibitem[Yamakawa et~al.(2011)Yamakawa, Namiki, and
  Ishikawa]{yamakawa2011dynamic}
Y.~Yamakawa, A.~Namiki, and M.~Ishikawa.
\newblock Dynamic manipulation of a cloth by high-speed robot system using
  high-speed visual feedback.
\newblock \emph{IFAC Proceedings Volumes}, 44\penalty0 (1):\penalty0
  8076--8081, 2011.

\bibitem[Shibata et~al.(2010)Shibata, Ohta, and Hirai]{shibata2010robotic}
M.~Shibata, T.~Ohta, and S.~Hirai.
\newblock Robotic unfolding of hemmed fabric using pinching slip motion.
\newblock In \emph{The Abstracts of the international conference on advanced
  mechatronics: toward evolutionary fusion of IT and mechatronics: ICAM
  2010.5}, pages 392--397. The Japan Society of Mechanical Engineers, 2010.

\bibitem[He et~al.(2016)He, Zhang, Ren, and Sun]{he2016deep}
K.~He, X.~Zhang, S.~Ren, and J.~Sun.
\newblock Deep residual learning for image recognition.
\newblock In \emph{Proceedings of the IEEE conference on computer vision and
  pattern recognition}, pages 770--778, 2016.

\bibitem[Bertiche et~al.(2020)Bertiche, Madadi, and
  Escalera]{bertiche2020cloth3d}
H.~Bertiche, M.~Madadi, and S.~Escalera.
\newblock Cloth3d: Clothed 3d humans.
\newblock In \emph{European Conference on Computer Vision}, pages 344--359.
  Springer, 2020.

\bibitem[Seita et~al.(2021)Seita, Florence, Tompson, Coumans, Sindhwani,
  Goldberg, and Zeng]{seita_bags_2021}
D.~Seita, P.~Florence, J.~Tompson, E.~Coumans, V.~Sindhwani, K.~Goldberg, and
  A.~Zeng.
\newblock {Learning to Rearrange Deformable Cables, Fabrics, and Bags with
  Goal-Conditioned Transporter Networks}.
\newblock In \emph{ICAR}, 2021.

\bibitem[Lillicrap et~al.(2015)Lillicrap, Hunt, Pritzel, Heess, Erez, Tassa,
  Silver, and Wierstra]{lillicrap2015continuous}
T.~P. Lillicrap, J.~J. Hunt, A.~Pritzel, N.~Heess, T.~Erez, Y.~Tassa,
  D.~Silver, and D.~Wierstra.
\newblock Continuous control with deep reinforcement learning.
\newblock \emph{arXiv preprint arXiv:1509.02971}, 2015.

\bibitem[Chi and Song(2021)]{chi2021garmentnets}
C.~Chi and S.~Song.
\newblock Garmentnets: Category-level pose estimation for garments via
  canonical space shape completion.
\newblock In \emph{The IEEE International Conference on Computer Vision
  (ICCV)}, 2021.

\end{thebibliography}
